\documentclass[sigconf]{acmart}

\AtBeginDocument{%
  \providecommand\BibTeX{{%
    \normalfont B\kern-0.5em{\scshape i\kern-0.25em b}\kern-0.8em\TeX}}}

\usepackage[T1]{fontenc}
\usepackage{lmodern}
\usepackage{multirow}
\usepackage{xcolor}
\usepackage{colortbl}

\begin{document}
\def\x{{\mathbf x}}
\def\L{{\cal L}}
\def\eg{\textit{e.g.}}
\def\ie{\textit{i.e.}}
\def\Eg{\textit{E.g.}}
\def\etal{\textit{et al.}}
\def\etc{\textit{etc}}
\def\P(#1){\Phelper#1|\relax\Pchoice(#1)}
\def\Phelper#1|#2\relax{\ifx\relax#2\relax\def\Pchoice{\Pone}\else\def\Pchoice{\Ptwo}\fi}
\def\Pone(#1){\Pr\left( #1 \right)}
\def\Ptwo(#1|#2){\Pr\left( #1 \mid #2 \right)}
\def\Pr{\mathbf{Pr}}
\newcommand{\blue}[1]{\textcolor{blue}{#1}}
\settopmatter{printacmref=false}
\title{\emph{I am empathetic and dutiful, and so will make a good salesman}: Characterizing Hirability via Personality and Behavior}

\author{Harshit Malik}
\affiliation{%
  \institution{Indian Institute of Technology Ropar}
}
\email{2017csb1078@iitrpr.ac.in}

\author{Hersh Dhillon}
\affiliation{%
  \institution{Indian Institute of Technology Ropar}
}
\email{2017csb1079@iitrpr.ac.in }

\author{Roland Goecke}
\affiliation{%
  \institution{University of Canberra}
}
\email{roland.goecke@canberra.edu.au}

\author{Ramanathan Subramanian}
\affiliation{%
  \institution{Indian Institute of Technology Ropar}
}
\email{s.ramanathan@iitrpr.ac.in}

%
%
%
%
%
%

\renewcommand{\shortauthors}{Trovato and Tobin, et al.}

\begin{abstract}
While personality traits have been extensively modeled as behavioral constructs, we model \textbf{\textit{job hirability}} as a \emph{personality construct}. On the {\emph{First Impressions Candidate Screening}} (FICS) dataset, we examine relationships among personality and hirability measures. Modeling hirability as a discrete/continuous variable with the \emph{big-five} personality traits as predictors, we utilize (a) apparent personality annotations, and (b) personality estimates obtained via audio, visual and textual cues for hirability prediction (HP). We also examine the efficacy of a two-step HP process involving (1) personality estimation from multimodal behavioral cues, followed by (2) HP from personality estimates.

Interesting results from experiments performed on $\approx$~5000 FICS videos are as follows. (1) For each of the \emph{text}, \emph{audio} and \emph{visual} modalities, HP via the above two-step process is more effective than directly predicting from behavioral cues. Superior results are achieved when hirability is modeled as a continuous vis-\'a-vis categorical variable. (2) Among visual cues, eye and bodily information achieve performance comparable to face cues for predicting personality and hirability. (3) Explanatory analyses reveal the impact of multimodal behavior on personality impressions; \eg, Conscientiousness impressions are impacted by the use of \emph{cuss words} (verbal behavior), and \emph{eye movements} (non-verbal behavior), confirming prior observations.  

\end{abstract} 




\maketitle

\section{Introduction}
Behavioral cues such as \emph{eye movement}, \emph{gestural}, \emph{facial, gazing} and \emph{cognitive patterns} have been employed in many human-centered applications such as mental and emotion state prediction~\cite{Parekh2018,Shukla2020}, personality trait estimation~\cite{Subramanian13,Batrinca12}, depression detection~\cite{Cummins2011}, privacy-preserving gender prediction~\cite{Bilalpur17} and cognitive load estimation~\cite{Bilalpur18,Lukanov16} over the past decade. Recently, there has been considerable interest in mining multimodal behavioral cues for predicting the outcome of \emph{job interviews}~\cite{NaimTGH18,Gucluturk2018,escalante2020modeling}. Given the large number of applications received by top companies on a daily basis~\cite{IBM_AI}, there has been an increased push for employing \emph{artificial hiring agents} (AHAs) to recruit candidates; the rationale is that AHAs assessing video resume\'s along with traditional ones can undertake recruitment in early stages, while trained recruiters focus their energies on assessing applicants' tangible and intangible skills in the later rounds.

In order to make the recruitment process transparent and fool-proof, AHAs need to \emph{justify} their decisions\footnote{alternatively, make recommendations to candidates} with \textit{sound reasoning}, termed \emph{\textbf{explainability}} in machine learning parlance. A handful of works have employed both verbal and non-verbal behavioral cues to predict a candidate's \emph{\textbf{hirability}}, \ie, the suitability of a candidate to be invited for interview later. Hirability prediction has been modeled either as a \emph{binary classification} (suitable/unsuitable) or \emph{regression} (measure of suitability, \eg, on a 1--5 scale) problem by these works. 

The general consensus among social psychologists is that \emph{personality traits} shape {human behavior} and influence a wide range of life outcomes; therefore, personality can conversely be viewed as a behavioral construct. We additionally model \emph{hirability as a personality construct}. While not positing this relation explicitly, hirability prediction (HP) works~\cite{Gucluturk2018,escalante2020modeling} have adopted the above rationale; \eg, authors of~\cite{escalante2020modeling} observe that the apparent personality trait annotations are highly predictive of hirability scores, with $R^2 = 0.91$ for a linear model. Likewise, an explanatory decision tree denoting (binary) hirability in terms of categorical big-five trait predictors is presented. A recent work~\cite{VinciarelliRDRA19} notes that one's \emph{empathy quotient} (EQ, denoting the drive to empathize) and \emph{systemizing quotient} (SQ, drive to analyze) significantly influence career choices; EQ is in turn associated with Extraversion and Agreeableness~\cite{Haas15}.  

\begin{figure}
\centering
  \includegraphics[width=\linewidth]{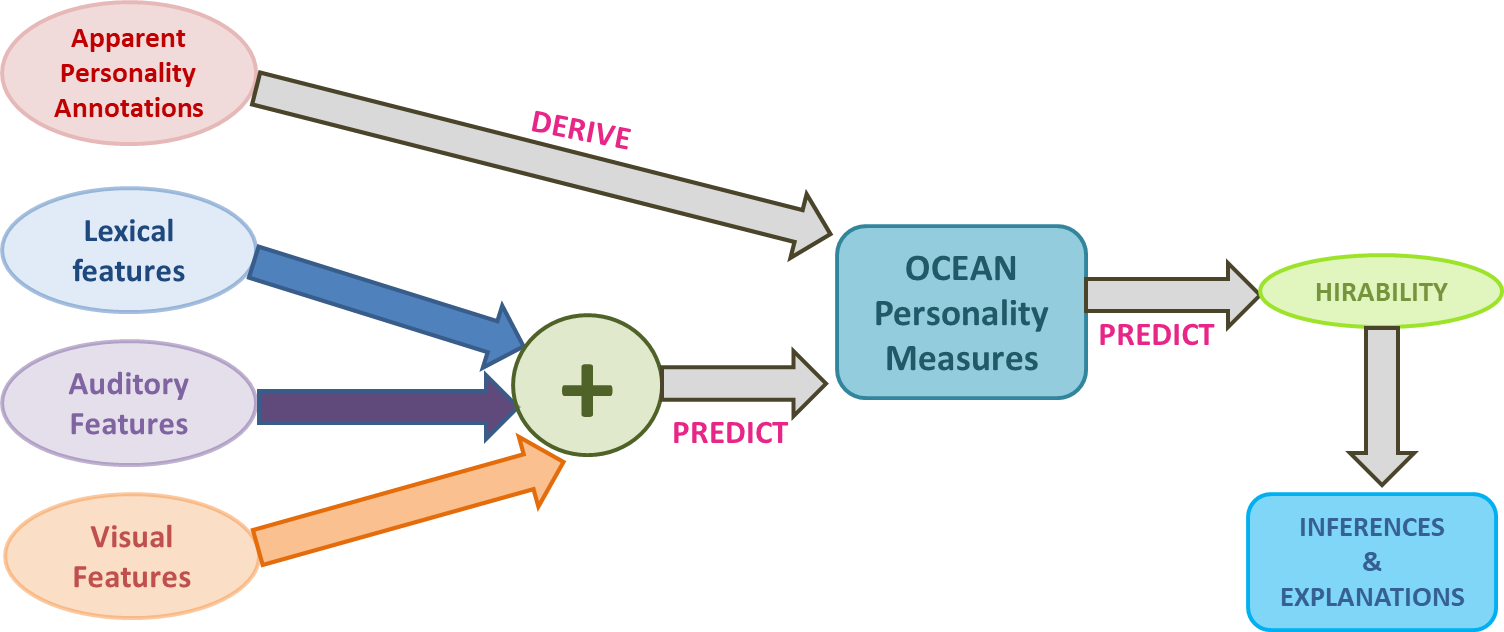}   
	\caption{Study Overview: We posit a significant correlation between a person's suitability for a vocation (termed \emph{hirability}) and his/her personality traits. To this end, OCEAN personality measures are either derived from first-impression annotations, or predicted from textual, auditory and visual behavioral cues. Continuous/categorical HP is then achieved from OCEAN measures, and explanations of both hirability and OCEAN personality predictions are attempted.} \vspace{-.5cm}
  \label{fig:moneyshot} 
\end{figure}

We posit hirability as a function of the Openness (O), Conscientiousness (C), Extraversion (E), Agreeableness (A) and Neuroticism (N) measures, also known as the \emph{big-five} personality traits. As in Fig.~\ref{fig:moneyshot}, we predict hirability (modeled as either a continuous/categorical variable) from behavioral cues as a two-step process: in the first step, OCEAN personality measures are either derived from manual ratings, or estimated from audio, visual and textual behavioral cues. HP from OCEAN measures is performed in the second step. Apart from facilitating explainability, we note that estimating hirability via this two-step process achieves better results than directly predicting from low-level behavioral cues. Overall, this work makes the following research contributions:
\begin{itemize}
\item[(1)] This work expressly explores the connections between \emph{personality traits} and \emph{hirability}. While one's personality may not directly determine his/her profession, prior works have noted the link between personality traits and career choices~\cite{VinciarelliRDRA19}. For recruiters, personality assessment in the early interview stages would help identify candidates who sync with the job requirements and company culture~\cite{Per-recruit}. While past works~\cite{Gucluturk2018,escalante2020modeling} have presented `proof-of-concept' results to show connections between personality and hirability annotations on the \textit{{First Impressions Candidate Screening}} dataset~\cite{escalante2020modeling}, we extensively explore relations between behavioral cues, personality traits and hirability.  
\item[(2)] With multiple modalities, we show that the two-stage hirability prediction framework performs \emph{better than} end-to-end HP from behavioral features. This is particularly surprising as end-to-end prediction is known to be less error-prone, and has fueled the use of deep neural networks for pattern recognition problems. 
\item[(3)] Further to (2), predicting hirability exclusively via the OCEAN personality traits enables better \emph{explainability} than a minimally interpretable `black-box' model involving high-dimensional behavioral inputs. 
\item[(4)] We specifically experimented with $\approx$~5000 FICS videos (4009 training~+~validation, 998 test) whose interview ratings are outside of the range [0.4,0.6], which can be construed as the `hirability gray area'. The primary objective behind this design is to \emph{explain} multimodal predictions relating to \emph{high}/\emph{low} hirability and apparent personality wherever possible. Explanatory analyses such as discovering the most informative word stems, and highlighting the most informative visual features reveal some interesting patterns; \eg, Impressions of Conscientiousness, which considerably influences hirability, are impacted by both \emph{cuss words} (verbal behavior) and \emph{eye movements} (non-verbal behavior), confirming prior findings~\cite{Con-swear,Hoppe18}.   

\end{itemize}

\begin{figure*}[!htbp]
\centering
  \includegraphics[width=0.32\linewidth]{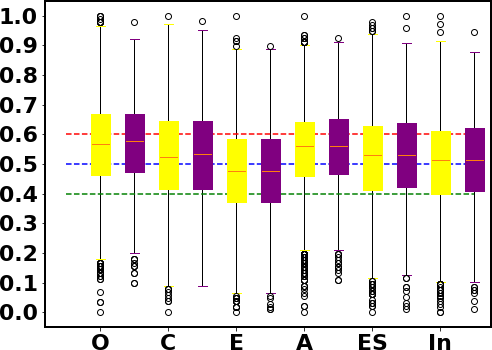}\hspace{0.1cm}\includegraphics[width=0.32\linewidth]{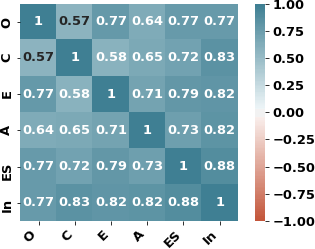}\hspace{0.1cm}\includegraphics[width=0.32\linewidth]{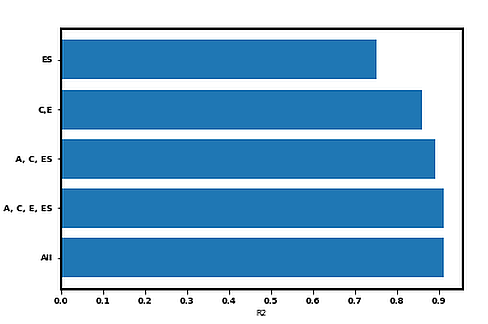}
	\centerline{(a)\hspace{0.33\linewidth}(b)\hspace{0.33\linewidth}(c)}\vspace{-.1cm}
  \caption{(a) Boxplots denoting distributions of the OCEAN \emph{personality trait} and the \emph{Interview} (In) measures from the \emph{First Impressions} dataset. \emph{Train} data (8000 videos) are depicted in yellow, and \emph{test} data (2000 videos) in purple. Inverse of the N trait is denoted as ES. (b) Heatmap depicting correlations among these six attributes. (c) $R^2$ values obtained for the \emph{best} linear regression model involving 1--5 personality trait predictors. Best viewed in color and under zoom.}\label{fig:FI_Summary1}
\vspace{.1cm}
  \includegraphics[width=0.49\linewidth, height=4cm]{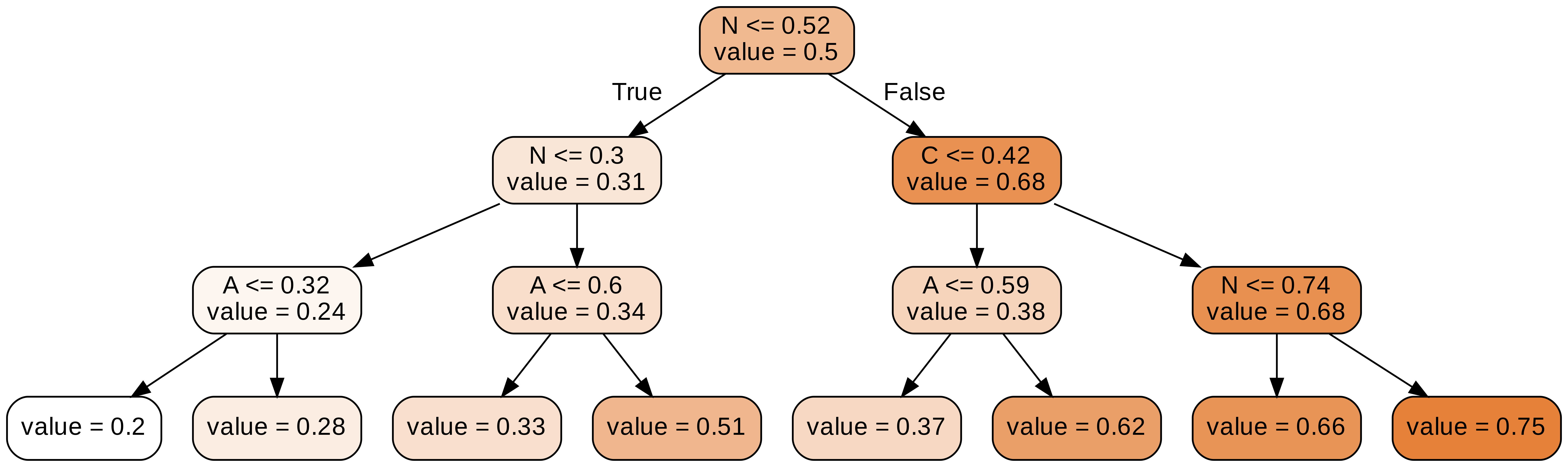}\hspace{0.12cm}\includegraphics[width=0.49\linewidth, height=4cm]{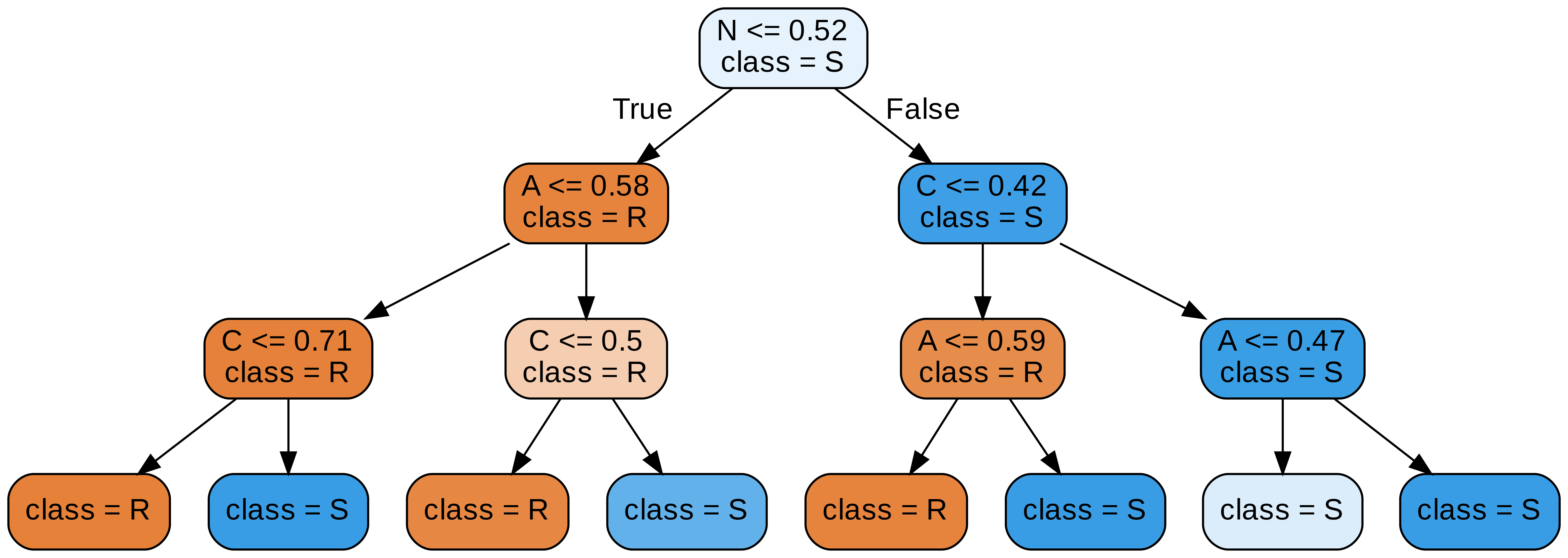}
  \caption{Decision trees obtained when hirability is modeled as a \emph{continuous} (left), and a \emph{categorical} (right) variable with the (S)elect and (R)eject classes for the \emph{sampled} FICS videos. I is estimated from continuous OCEAN scores in both cases.} 
	\label{fig:FI_Summary2}
	\vspace{-.4cm}
\end{figure*}

\section{Related Work}
We expressly focus on (a) trait estimation from behavior, and (b) explainable HP, while performing the literature survey.

\subsection{Trait prediction from behavioral cues}
For long, there has been consensus among social psychologists that \emph{personality traits} shape human behavior, and influence a large number of our life outcomes. Therefore, design of human-centered intelligent systems has primarily focused on the inverse problem; that of employing \emph{behavioral cues} (typically audio and visual) to deduce attributes such as the big-five personality traits~\cite{Subramanian2010, Subramanian13,Hoppe18}, and traits highly correlated with personality such as \emph{depression}~\cite{Cummins2011} and \emph{stress}~\cite{Finnerty16}. 

Recently, hirability prediction (HP) has been attempted by a number of researchers~\cite{NaimTGH18,Gucluturk2018,escalante2020modeling} from multimodal behavioral cues. Fool-proof HP would enable large organizations to employ artificial hiring agents (AHAs) and effectively reach out to the vast number of applicants contacting them on a daily basis. HP algorithms have typically modeled hirability (or interview variable I) as an adjunct to the OCEAN personality traits, thereby predicting IOCEAN traits from multimodal behavior. However, we posit a strong connect between positional requirements and personality traits, as recruiters would typically look for certain traits in candidates reflective of the organization's culture and values. Moreover, recent studies~\cite{VinciarelliRDRA19} have proposed a connection between the \emph{empathy quotient} psychometric, relating to Conscientiousness and Agreeableness, and one's career choices. Given these recent findings, this work explicitly explores the connection between hirability and the big-five OCEAN personality traits. Our experiments show that estimating hirability from OCEAN measures is more effective than HP performed directly from multimodal behavior.        

\subsection{Explainable hirability prediction}
To ensure transparent and fool-proof recruitment, AHAs need to be capable of \emph{justifying} their decisions/recommendations, termed \emph{explainability} in machine learning parlance. The handful of works that have examined HP from behavior have essentially focused on isolating behavioral correlates of the IOCEAN traits from quantitative results. Two recent works on HP~\cite{Gucluturk2018,escalante2020modeling} have loosely explored \emph{explainability in HP}. Specifically,~\cite{Gucluturk2018} explains hirability predictions based on personality \emph{annotations}, while~\cite{escalante2020modeling} shows \emph{typical faces} reflective of apparent traits. Differently, we show (a) how candidates' verbal behavior impacts their apparent traits, and (b) what deep neural networks focus on, given the candidate's face image or their portrait with the face blurred, by way of explaining trait predictions.     

\subsection{Inferences from literature survey}
Summarizing prior HP works, we note that (1) HP has been attempted from behavioral measures, but not from personality estimates obtained via behavioral measures; we posit a strong correlation between personality and hirability measures, and hypothesize that HP would be more effective and explainable if modeled as an exclusive function of the OCEAN traits; the effectiveness of performing HP from OCEAN measures is confirmed by our experiments. (2) Very limited material pertaining to \emph{explainability} of IOCEAN impressions is available; We explicitly perform explanative analyses to show how language and visual cues influence trait impressions, particularly the Conscientiousness trait. 


\section{Overview of the FICS dataset}\label{FICS_overview}

\begin{figure}[b]
\includegraphics[width=\linewidth]{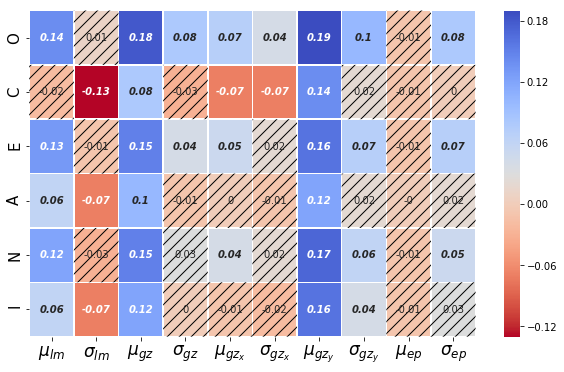}\vspace{-.3cm}
 \caption{Pearson correlations between visual behavioral measures and IOCEAN annotations for 4009 \emph{sampled} FICS training videos. Non-significant correlations are crossed out.}\label{fig:FI_Summary3}\vspace{-.4cm}
\end{figure}

This section is designed to provide readers with an overview of the First Impressions Candidate Screening (FICS)dataset, and serves as a prelude for the forthcoming sections. Interested readers may refer to~\cite{escalante2020modeling,Gucluturk2018} for further details. 

The FICS video dataset comprises 10000 videos (6000 training, 2000 validation and 2000 testing), and was designed with the objective of developing AHAs to make decisions/recommendations based on multimedia CVs~\cite{escalante2020modeling}. All videos contain labels for \emph{apparent} OCEAN personality traits (reflecting first impressions of a human observer viewing the CV), and a \emph{hirability}/\emph{interview} trait, indicating whether the video candidate should be invited for a job interview. OCEAN and interview (I) scores are denoted within the [0,1] range. Since Neuroticism (N) is a negative trait, N scores are replaced by Emotional Stability (ES) scores in FICS, and therefore, the terms N and ES are used interchangeably from hereon even if they strictly refer to opposite traits.    

Fig.~\ref{fig:FI_Summary1}(a) shows the FICS rating distributions. For experimental purposes, we combined the training$+$validation videos (8000 in total). Roughly similar training and test distributions can be noted for the I, C and E traits from Fig.~\ref{fig:FI_Summary1}(a). Annotation distributions for all traits are roughly Gaussian, and 70\% of A scores fall within one standard deviation from the mean implying a `tight' clustering, while the `loosest' clustering is noted for the ES trait, with 67\% samples falling within the same range. In terms of inter-quartile range (IQR) denoting the difference between the $75^{th}$ and $25^{th}$ percentiles, A has the lowest IQR of 0.18, while C has the highest IQR of 0.22. I score has an IQR of 0.21, and can be seen to be highly correlated with the OCEAN traits from Fig.~\ref{fig:FI_Summary1}(b). Finally, a linear regression model with OCEAN measures predicting the I score (Fig.~\ref{fig:FI_Summary1}(c)) shows N as the single-best predictor of I scores, and O as the worst. 

We set out to explore if \emph{hirability} predictions could be explained, at least for the \emph{high} and \emph{low} hirability samples, and therefore, we ignored videos with I scores between [0.4,0.6] for our analysis. All our experiments are therefore performed on the \emph{sampled FICS dataset}, with 4009 training samples (2134 +ve and 1875 -ve), and 998 test samples (544 +ve and 444 -ve), aggregating 5007 out of the 10K original videos. 

In subsequent sections, we present results where the I trait is modeled as a \emph{continuous}/\emph{categorical} variable, and the OCEAN predictor variables are also modeled as \emph{continuous}/\emph{categorical}. We predict both the I and OCEAN scores from multimodal behavioral measures, and show that estimating I from OCEAN estimates is more effective than direct prediction from behavioral cues. To this end, we define the regression and classification {performance metrics} as the I score \textit{estimation accuracy}, $\text{Acc} = 1-\text{MAE}$, where MAE denotes the mean absolute error over the test set; this evaluaton metric is also employed in~\cite{Gucluturk2018}. 

Fig.~\ref{fig:FI_Summary2} illustrates the computed decision trees when I is predicted as a continuous and as a categorical variable from OCEAN annotations. Both decision trees obtained for the sampled FICS dataset place minimal emphasis on the O trait as an I predictor, similar to the linear regression model in Fig.~\ref{fig:FI_Summary1}(c). The regression and classification decision trees achieve an accuracy of 0.96 and 0.99 respectively on the sampled FICS test set. We will compare all models presented in the next section with the above \emph{annotation-based benchmarks}.

To provide a flavor of how behavioral measures affect I and OCEAN impressions, Fig.~\ref{fig:FI_Summary3} presents correlations among visual behavioral cues and the IOCEAN traits based on \emph{Openface}~\cite{Balt18} outputs. FICS videos are $\approx$ 15s long, and upon dividing each video into non-overlapping 1s \emph{thin slices}~\cite{Subramanian13}, we computed $\mu, \sigma$ statistics for: motion of 68 facial landmarks ($\mu_{lm}, \sigma_{lm}$), gaze direction vector ($\mu_{gz}, \sigma_{gz}$), head pan ($\mu_{gz_x}, \sigma_{gz_x}$), head tilt ($\mu_{gz_y}, \sigma_{gz_y}$), and the proportion of time for which the candidates' eyes are pointing towards the camera/viewer ($\mu_{ep}, \sigma_{ep}$), over all 1s thin slices.  

Focusing only on significant correlations > 0.1 in magnitude, one can observe the following: (a) consistently high facial landmark movement over all thin-slices is +vely correlated with the O, E and ES traits (\emph{highly expressive candidates are open-minded, extroverted and emotionally balanced}); (b) consistently high head-tilt motion is +vely correlated with all traits (\emph{head nodding is viewed as +ve behavior}), and (c) intermittent landmark motion, captured by high $\sigma_{lm}$, is -vely correlated with Conscientiousness (\emph{unprepared candidates are more likely to exhibit awkward/sudden facial movements}). Evidently, IOCEAN annotations can be explained by visually examining the candidate's non-verbal behavior in a fine-grained manner.


\section{Behavioral Cues to Hirability} 

This section examines (a) the utility of various language (verbal), auditory and visual cues for HP, (b) compares HP from behavioral cues vis-\'a-vis the two-step process of personality estimation from behavior, followed by HP from OCEAN estimates, and (c) attempts to explain prediction patterns relating to personality and hirability.

\subsection{Verbal (Textual) Cues}
As the FICS dataset is accompanied by transcriptions of the candidate videos~\cite{escalante2020modeling,Gucluturk2018}, we examined the impact of candidates' language on their apparent OCEAN and I scores via 4009 {training}~videos and 998 test videos. 

\subsubsection{Experimental Settings}
Videos having I score $\leq 0.4$ considered as -ve samples, while videos with $I\geq 0.6$ were considered as +ve samples for classification. For both continuous (regression) and categorical prediction (classification) of I scores, \emph{continuous} OCEAN estimates derived from textual cues were used. The following feature extraction and regressor/classifier frameworks were examined. \\

\noindent{\textbf{Bag of Words (BoW) feature extraction:}} As in~\cite{Gucluturk2018}, we adopted the BoW approach for text analyses. From video transcripts, stopwords were removed and we used 4 word categories :- adjectives, adverbs, verbs and nouns to construct our vocabulary. The top 5000 most frequently appearing words were selected as feature vectors; each transcript is therefore denoted by a 5000-D vector, which was input to the following algorithms. \\

\noindent{\textbf{Regressor/classifier frameworks:}} For regression, we employed the random forest (RF) and Support vector Regressor (SVR), while for classification, we used the (a) Naive Bayes (NB) classifier provided by NLTK (\url{https://www.nltk.org/}), (b) Binomial Naive Bayes (B-NB),  (c) Logistic Regression (LR), (d) Support Vector Regressor (SVR) and (e) AWD-LSTM, the stochastic gradient descent-based long short-term memory pipeline provided by FastAI (\url{https://www.fast.ai/}). 

%


\subsubsection{Quantitative and Qualitative Results}\label{sec:quantqual}
Table~\ref{tab: RegClass_text} presents regression and classification on the IOCEAN traits. Table~\ref{tab: Inter_Pers_text} presents continuous/categorical I score prediction from regression-based OCEAN estimates in Table~\ref{tab: RegClass_text}. Furthermore, we found the top 10 \emph{most informative} word stems for each trait via the NB classifier (Table~\ref{tab:Qual_text}). Informative word stems were identified as follows: We computed the relative likelihood of selection (\emph{S}) vs rejection (\emph{R}) given \emph{stem} via importance weights (IW) as $\text{IW}_{stem} =$ {$\P(S \mid stem)$}/{$\P(R \mid stem)$}. Therefore, $\text{IW}_{stem} = 10$ implies that the selection likelihood of a candidate using the word \emph{stem} is 10 times \emph{higher} than one who does not use the \emph{stem}; In short, \emph{stems with +ve IWs positively impact trait impressions, while stems with -ve IWs elicit a negative trait impression in the observer}. +ve and -ve stems corresponding to the IOCEAN traits are respectively coded in green and red in Table~\ref{tab:Qual_text}.

\begin{table}[t]
\scriptsize
\caption{Quantitative IOCEAN prediction from textual cues.}\label{tab: RegClass_text}\vspace{-.3cm}
\begin{center}
\begin{tabular}{|c|cccccc|}
\hline
\textbf{Model} & \textbf{I} & \textbf{O} & \textbf{C} & \textbf{E} & \textbf{A} & \textbf{N} \\
\hline
\multicolumn{7}{|c|}{\textbf{Regression}} \\
\hline
\textbf{RF} & 0.817 &  0.852 & 0.839 & 0.840 & 0.858 & 0.832 \\
\textbf{SVR} & 0.837 &  0.863 & 0.853 & 0.851 & \textbf{0.868} & 0.848 \\
\hline

\multicolumn{7}{|c|}{\textbf{Classification}} \\
\hline
\textbf{NB} & 0.632 &  0.618 & 0.638 & 0.605 & 0.617 & 0.639 \\
\textbf{B-NB} & 0.620 & 0.650 & 0.618 & 0.602 & 0.605 & 0.653 \\
\textbf{LC} & 0.606 & 0.610 & 0.602 & 0.592 & 0.578 & 0.643 \\
\textbf{SVC} & \textbf{0.665} & 0.651 & 0.664 & 0.631 & 0.647 & 0.676 \\
\textbf{AWD-LSTM} & 0.624 &  0.627 & 0.634 & 0.612 & 0.624 & 0.637 \\
\hline
\end{tabular}
\vspace{0.1cm}
\caption{HP from OCEAN measures ({textual} cues).}\label{tab: Inter_Pers_text}\vspace{-.2cm}
\begin{tabular}{|c|c c c|}
\hline
\textbf{Model} & \textbf{RF} & \textbf{SVR} & \textbf{SVC} \\
\textbf{Regression} & 0.847 & \textbf{0.849} & - \\
\textbf{Classification}  - & 0.652 & &\textbf{0.657} \\ \hline
\end{tabular}
\end{center}
\vspace{.1cm}
\caption{Exemplar +ve (green) and -ve (red) word stems for the IOCEAN traits. IWs specified in brackets.}\label{tab:Qual_text}\vspace{-.2cm}
\begin{center}
\begin{tabular}{|c|c c c c|}
\hline
\textbf{{I}} & \cellcolor{green!25} \emph{lucki} (6.7) & \cellcolor{red!25}\emph{dead} (-6.5) & \cellcolor{green!25}{\emph{achiev} (6.1)} & \cellcolor{green!25}{\emph{discuss} (6.1)} \\ 
\textbf{O}    &  \cellcolor{red!25}{\emph{perfectli} (-6.6)} & \cellcolor{green!25}{\emph{limit} (6.6)} & \cellcolor{green!25}{\emph{young} (6.6)} &  \cellcolor{red!25}{\emph{knowledg} (-5.8)} \\
\textbf{C}    &  \cellcolor{red!25}{\emph{fuck} (-13.0)} & \cellcolor{green!25}{\emph{healthy} (10.6)} & \cellcolor{green!25}{\emph{diet} (8.2)} &  \cellcolor{red!25}{\emph{dead} (-6.2)} \\
\textbf{E} & \cellcolor{red!25}{\emph{address} (-7.4)} & \cellcolor{green!25}{\emph{discuss} (6.6)} & \cellcolor{green!25}{\emph{hobbi} (6.0)} & \cellcolor{green!25}{\emph{fashion} (6.0)} \\
\textbf{A} & \cellcolor{green!25}{\emph{mention} (8.2)} & \cellcolor{green!25}{\emph{discuss} (6.1)} & \cellcolor{red!25}{\emph{maintain} (-5.8)} & \cellcolor{green!25}{\emph{monitor} (5.5)} \\ 
\textbf{ES(N)} & \cellcolor{green!25}{\emph{lucki} (6.7)} & \cellcolor{red!25}{\emph{dead} (-6.5)} & \cellcolor{green!25}{\emph{discuss} (6.1)} & \cellcolor{red!25}{\emph{maintain} (-5.8)} \\ 
\hline
\end{tabular}
\end{center}
\vspace{-.3cm}
\end{table} 

\subsubsection{Discussion} From Tables~\ref{tab: RegClass_text},\ref{tab: Inter_Pers_text},\ref{tab:Qual_text}, we make the following remarks. (a) Continuous IOCEAN estimates are more effectively predicted by all models as compared to categorical values, as per the Acc values for regression and classification in Table~\ref{tab: RegClass_text}. (b) Continuous I score prediction from textual features (max Acc of 0.837) is less effective than predicting from estimated OCEAN measures (max Acc = 0.849) as per Tables~~\ref{tab: RegClass_text} and~\ref{tab: Inter_Pers_text}. (c) Most importantly, \emph{\textbf{intuitive connections}} between {word stems }and {traits }are noted via IWs. \emph{E.g.} stems such as \emph{achieve}, \emph{lucki} and \emph{discuss} are seen as +ve with respect to hirability, while \emph{dead} is deemed -ve. Use of \emph{dead} is also as a sign of anxiety, conveying high Neuroticism. 

The word \emph{discuss} is seen as +ve in the context of Agreeableness and Extraversion, while \emph{hobbi} and \emph{fashion} also convey an impression of high Extraversion consistent with Ashton's theory that extraverts engage in \emph{attractive} social activities~\cite{Ashton2002}. Conscientiousness impressions, characterized by diligence and uprightness, are negatively impacted by the use of \emph{cuss words}~\cite{Con-swear}, and positively impacted by the use of words such as \emph{healthy} and \emph{diet} related to well-being. Overall, while examining verbal behavior requires the generation of transcripts which is tedious/challenging, our experiments reveal the utility of such an exercise, as word choices impact both \emph{trait} and \emph{hirability} impressions. 
  
\begin{table}[!tbp]
\scriptsize
\renewcommand{\arraystretch}{1.2}
\caption{Description of extracted audio features.}\label{tab:audio_feat}
\vspace{-.4cm}
\begin{center}
\begin{tabular} { |m{0.17\linewidth}|m{0.80\linewidth}|}
 \hline
 \textbf{MFCCs} & Form a representation where frequency bands are not linear but distributed on the mel-scale \\
 \textbf{Energy} & Squared-sum of signal values, normalized by the frame length \\
\textbf{ZCR} & Zero crossing rate of the signal within a particular frame\\
\textbf{Tempo} & Beats per minute \\
\textbf{Sp. flatness} & Measure to quantify \emph{noise-like} trait of a sound spectrum\\
\textbf{Sp. bandwidth} & $p$'th-order spectral bandwidth, default $p = 2$ \\
\textbf{Sp. roll-off} & Frequency below which 90\% spectrum is concentrated \\
\textbf{Sp. contrast} & For each sub-band, compare mean energy of top quantile with mean of bottom quantile.\\
\textbf{Tonnetz} & Tonal centroid features \\
\hline
\end{tabular}
\end{center}
\vspace{-.5cm}
\end{table}

\subsection{Auditory cues}
\subsubsection{Feature extraction}\label{sec:features}
For predicting IOCEAN traits from audio, we extracted low-level speech signal statistics from the \emph{Librosa} library (\url{https://librosa.github.io/librosa/feature.html}), and audio spectrograms. \emph{Librosa} features were fed to a random forest (RF), while speech sprectrograms were fed to a VGG11 (CNN) for regression/classification as in Table~\ref{tab:audio_IOCEAN}. A total of 56 audio statistics including $\mu, \sigma$ for 20 MFCC coefficients (Table~\ref{tab:audio_feat}) were employed for analysis. 

\subsubsection{Experimental Settings}
\begin{figure}[!htbp]
\includegraphics[width=0.9\linewidth]{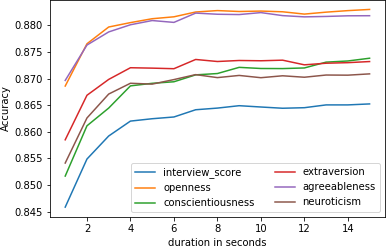}\vspace{-.2cm}
\caption{\label{fig:audio_dur} IOCEAN prediction from \emph{Librosa} features with varying time windows.}\vspace{-.3cm}
\end{figure}

\begin{sloppypar}
For IOCEAN estimation, we considered the IOCEAN traits as both continuous and categorical; regression and classification results are coded as (R) and (C) respectively in Table~\ref{tab:audio_IOCEAN}. As a second step, we predicted continuous/categorical I scores from continuous/discrete OCEAN estimates (Table~\ref{HP_Ocean_audio}). We also adopted the \emph{thin-slice} approach (as in Sec.~\ref{FICS_overview}) for audio analysis, aggregating 1s \emph{Librosa} statistics over 2--15 second time-windows to predict continuous IOCEAN measures~(Fig.~\ref{fig:audio_dur}). Results in Table~\ref{tab:audio_IOCEAN},\ref{HP_Ocean_audio} correspond to 15s time windows (equal to the length of FICS videos).
\end{sloppypar}

\begin{table}[!htbp]
\fontsize{7.5}{7.5}\selectfont
\caption{Audio performance for IOCEAN estimation.} \label{tab:audio_IOCEAN}\vspace{-.2cm}
\begin{center}
\begin{tabular}{|c|cccccc|}
\hline
\textbf{Model} & \textbf{I} & \textbf{O} & \textbf{C} & \textbf{E} & \textbf{A} & \textbf{N} \\
\hline
\textbf{RF (R)} & {0.8783} & \textbf{0.8916} & {0.8863} & {0.8819} & {0.8901} & {0.8803}\\
\textbf{CNN (R)} & 0.8799 & 0.8835 & 0.8809 & 0.8773 & 0.8880 & 0.8768 \\
\hline
\textbf{RF (C)} & \textbf{0.8116} & {0.7766} & {0.8036} & {0.7786} & {0.7725} & {0.8066}\\
\textbf{CNN (C)} & 0.7565 & 0.7455 & 0.7345 & 0.7445 & 0.7385 & 0.7615 \\
\hline
\end{tabular}

\vspace{0.1cm}
\caption{HP from OCEAN measures (audio cues). Labels R and C denote continuous/categorical OCEAN estimates.}\label{HP_Ocean_audio}\vspace{-.2cm}
\begin{tabular}{|cc|cccc|}
\hline
\multicolumn{2}{|c|}{\textbf{Regression}} & \multicolumn{4}{c|}{\textbf{Classification}}\\
 \hline
\textbf{RF (R)} & \textbf{CNN (R)} & \textbf{RF (R)} & \textbf{CNN (R)} & \textbf{RF (C)} & \textbf{CNN (C)}\\
\hline
\textbf{0.8946} & 0.8821 & 0.8156 & 0.7876 & \textbf{0.8235} & 0.7912 \\
\hline
\end{tabular}
\end{center}
\vspace{-.2cm}
\end{table}

\subsubsection{Results and Discussion}
We make the following remarks from our experimental results. (1) As with text analysis, continuous IOCEAN prediction is better achieved (max Acc = 0.8916) than discrete (max Acc = 0.8116). (2) Consistent with text-based results, better prediction of I scores is achieved from continuous OCEAN estimates (max Acc = 0.8946), than from audio features (max Acc = 0.8799). (3) The time-window varying experiment was designed to verify if 15s of audio data is indeed necessary for accurate IOCEAN prediction. From Fig.~\ref{fig:audio_dur}, we note that the Acc results saturate beyond 6s, reflecting that reliable trait estimation is achievable upon observing only \emph{tiny} behavioral episodes, and conveying that 15s windows is redundant for audio-based trait estimation. Overall, the O and A traits are best reflected by audio features, while Interview scores are not well predicted via \emph{Librosa} statistics.        

\subsection{Visual Analysis}\label{sec:vid_mod}
Non-verbal behavior cues, especially visual, have been extensively employed for human-centered applications earlier~\cite{Subramanian2010,Cummins2011,Subramanian13,Finnerty16}. This is due to the fact that visual behaviors such as gazing, facial emotions and movements, and body movements convey a significant amount of informative and communicative cues during social interactions. Especially during interview sessions, visual behaviors can convey a lot of information (\emph{is the candidate calm or emotional when facing a tough situation?}) to the interviewer.

\begin{figure}[!htpb]
\centering
\includegraphics[width=0.32\linewidth,height=2.3cm]{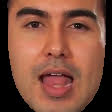}\hspace{0.1cm}\includegraphics[width=0.18\linewidth]{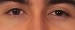}\hspace{0.1cm}\includegraphics[width=0.35\linewidth,height=2.3cm]{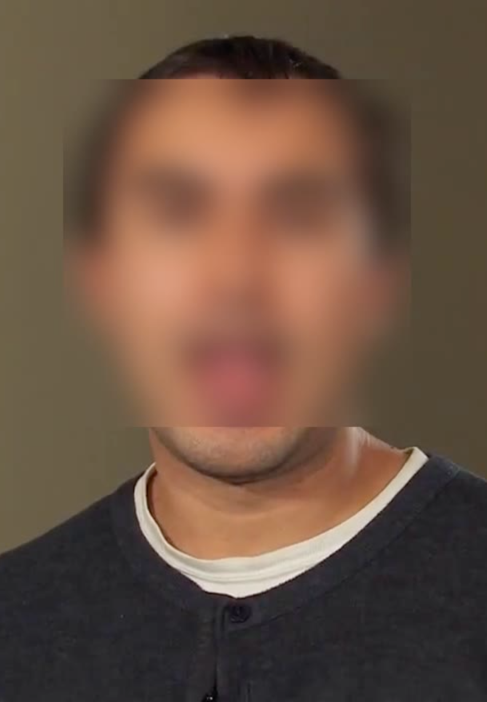}
\caption{\label{fig:vis_cues} Inputs to the visual model include the cropped face image (left), cropped eye region (center) and face-blurred portrait to examine the influence of holistic body movements for trait prediction.}\vspace{-.3cm}
\end{figure}

Given the critical contribution of visual behavior to IOCEAN prediction, we opted to examine \emph{multiple} visual cues different from prior HP works~\cite{escalante2020modeling,Gucluturk2018,NaimTGH18}. Instead of examining only facial cues for trait prediction, we also proceeded to examine the \emph{eye} and the \emph{body movements}; we therefore additionally input an eye-crop and a body-crop with the face blurred (Fig.~\ref{fig:vis_cues}) to the prediction frameworks, to evaluate the contribution of eye and body movements towards IOCEAN prediction. The face and eye-crops are obtained via \emph{Openface}~\cite{Balt18}, while the face-blurred body-crop is obtained by smoothing the facial region in the video frame using a Gaussian filter, so that the facial details are not apparent to the observer.

\subsubsection{Experimental settings}
We considered the following prediction models in our experiments. \\

\noindent \textbf{2D-CNN:} A 19 layered VGG model, which processes 2D frame information was used. The VGG output layer was removed, and two hidden fully-connected layers with 512 and 64 neurons respectively were added along with output layer involving 6 neurons (one neuron each for the IOCEAN traits). Mean squared error (MSE) for regression, and binary cross-entropy (BCE) loss for classification were used during training on a single, representative frame from the video sequence, with learning rate of 1e-4 and a batch size of 64. \\

\noindent \textbf{3D-CNN:} An 18 layered ResNet-3D model (\url{https://arxiv.org/abs/1711.11248}), with pre-trained weights for human activity recognition, was used. The 3D-CNN model took inputs from 16 uniformly spaced visual frames, sampled at $t=0,1,\ldots,15$ seconds into the video. The ResNet-3D output layer was removed, and two hidden fully-connected layers with 128 and 32 neurons respectively were added instead, along with final output layer of 6 neurons. The 16 stacked frames are re-sized to $112x112$ prior to input. Mean squared error (MSE) Loss was used during training (3D-CNN was employed only for regression), with learning rate 1e-4 and batch size 32. \\

\noindent \textbf{LRCN:} which denotes a Long-term Recurrent Convolutional Neural network (\url{https://arxiv.org/abs/1411.4389}) with a pre-trained ResNet-50 encoder and a single-layer LSTM decoder. This model takes 40 uniformly-spaced video frames as input; the encoder CNN learns 512-D features for each frame, which are fed to the LSTM decoder across different time frames. The 512-D LSTM output is fed into a linear layer of size 256, which is then connected to the final layer composed of 6 neurons. L1-loss was used for model training, with learning rate for the pre-trained ResNet set to 1e-6, and varying between 1e-4 to 1e-5 for other layers. The Adam optimizer was used to train the LRCN.

\begin{table}[!htbp]
\fontsize{7}{7}\selectfont
\caption{IOCEAN Regression from visual cues: 2D, 3D and LRC refer to 2D-CNN, 3D-CNN and LRCNN. Codes F, E and B denote facial, eye and body cues.}\label{tab:IOCEAN_video_reg} \vspace{-.3cm}
\begin{center}
\begin{tabular}{|ccccccccc|}
\cline{2-9} \multicolumn{1}{c|}{}  & \textbf{2D(F)} & \textbf{2D(B)} & \textbf{2D(E)} & \textbf{3D(F)} & \textbf{3D(B)} & \textbf{LR(F)} & \textbf{LR(B)} & \textbf{LR(E)}\\
\hline
\textbf{I} & 0.897 & 0.909 & 0.869 & \textbf{0.910} & 0.903 & 0.902 & 0.896 & 0.891 \\
\textbf{O} & 0.897 & 0.903 & 0.881 & {0.903} & 0.903 & 0.896 & 0.896 & 0.887 \\ 
\textbf{C} & 0.895 & {0.909} & 0.876 & 0.904 & 0.900 & 0.901 & 0.894 & 0.889 \\
\textbf{E} & 0.894 & 0.898 & 0.874 & {0.908} & 0.897 & 0.895 & 0.892 & 0.886 \\
\textbf{A} & 0.893 & 0.902 & 0.878 & {0.904} & 0.902 & 0.900 & 0.896 & 0.892 \\
\textbf{N} & 0.889 & 0.896 & 0.867 & {0.902} & 0.895 & 0.892 & 0.885 & 0.882 \\
\hline
\end{tabular}
\end{center}
\vspace{0.1cm}
\caption{IOCEAN Classification from visual cues: 2D, 3D and LRC refer to 2D-CNN, 3D-CNN and LRCNN respectively. Codes F, E and B denote facial, eye and body cues.}\label{tab:IOCEAN_video_class} \vspace{-.3cm}
\begin{center}
\begin{tabular}{|cccccc|}
\cline{2-6} \multicolumn{1}{c|}{}  & \textbf{2DC (F)} & \textbf{2DC (B)} & \textbf{2DC (E)} & \textbf{LRC (F)} &\textbf{LRC (E)} \\
\hline
\textbf{I} & 0.7856 & {0.8287} & 0.7101 & 0.8106 & 0.7934 \\
\textbf{O} & 0.7525 & {0.7826} & 0.7011 & 0.7675 & 0.7512 \\
\textbf{C} & 0.7776 & {0.8267} & 0.7101 & 0.7996 & 0.7853 \\
\textbf{E} & 0.7545 & 0.7745 & 0.6911 & {0.7916} & 0.7733 \\
\textbf{A} & 0.7295 & {0.7796} & 0.6670 & 0.7605 & 0.7442 \\
\textbf{N} & 0.7766 & \textbf{0.8307} & 0.7081 & 0.8036 & 0.7944 \\
\hline
\end{tabular}
\end{center}
\vspace{0.1cm}
\caption{HP from \emph{continuous} OCEAN estimates. 2D, 3D and LR refer to 2D-CNN, 3D-CNN and LRCNN. F, E and B codes in brackets stand for facial, eye and body cues. R/C codes denote continuous/categorical HP.}\label{tab:I_OCEAN_con_video} \vspace{-.3cm}
\begin{center}
\begin{tabular}{|m{0.7cm}m{0.7cm}m{0.7cm}m{0.7cm}m{0.7cm}m{0.7cm}m{0.7cm}m{0.7cm}|}
\hline
\textbf{2D(FR)} & \textbf{2D(BR)} & \textbf{2D(ER)} & \textbf{3D(FR)} & \textbf{3D(BR)} & \textbf{LR(FR)} & \textbf{LR(BR)} & \textbf{LR(ER)}\\
\hline
\hline
0.90 & 0.91 & 0.87 & \textbf{0.91} & 0.91 & 0.91 & 0.91 & 0.90 \\
\hline\hline
\textbf{2D(FC)} & \textbf{2D(BC)} & \textbf{2D(EC)} & \textbf{3D(FC)} & \textbf{3D(BC)} & \textbf{LR(FC)} & \textbf{LR(BC)} & \textbf{LR(EC)}\\\hline
0.82 & 0.85 & 0.74 & \textbf{0.87} & 0.86 & 0.85 & 0.84 & 0.83 \\
\hline
\end{tabular}
\end{center}
\vspace{-0.2cm}
\end{table}

\subsubsection{Results \& Discussion} Tables~\ref{tab:IOCEAN_video_reg},~\ref{tab:IOCEAN_video_class} and~\ref{tab:I_OCEAN_con_video} present trait predictions from the multiple visual cues. From Table~\ref{tab:IOCEAN_video_reg}, which estimates continuous IOCEAN values from the face, eye and body cues, we make the following remarks: (1) In terms of the general predictive power, the 3D-CNN is more potent than the 2D-CNN and LRCNN frameworks. Acc values $\geq 0.9$ are often observed with the 3D-CNN, while the 2D-CNN and LRCNN perform slightly inferiorly. (2) An interesting finding is that the eye and body-cues achieve performance \emph{comparable} to the face cue. This is particularly important as it opens up the possibility of AHAs being able to examine video CVs and make reasonable trait-related decisions \textbf{\emph{while honoring the candidate's privacy}} (processing only a mid-to-low resolution image of the eye, or blurring the face will render the facial information unusable as a biometric). (4) The face cue is nevertheless critical, and produces the best prediction for the Interview trait. (5) Among OCEAN traits, C and A are the two best-predicted traits from visual cues.

Focusing on IOCEAN classification results in Table~\ref{tab:IOCEAN_video_class}, in line with the text and audio-based results, considerably lower Acc values than regression are noted for classification. Interestingly, body cues produce the best categorical IOCEAN estimates, and achieve considerably better performance than face or eye cues. This results indicates that perhaps, \textbf{\emph{a fine-grained visual examination of the candidate's behavior may not be necessary to make a coarse-grained decision (\ie, suitable or unsuitable) regarding the candidate's hirability}}. A distant examination could still be adequate. Among IOCEAN traits, N is predicted best based on body cues by the 2D-CNN, which is revealing as the N trait is associated with anxiety, which may manifest via body-fidgeting, \etc.  

Examining Table~\ref{tab:I_OCEAN_con_video} which presents continuous/categorical HP from OCEAN estimates, we again note that $\text{Acc}\geq 0.9$ is achieved for all conditions (second table row), except with the 2D-CNN employing eye information. The best prediction of categorical I labels (Acc = 0.87) is achieved when continuous OCEAN scores are estimated employing facial information; this implies that \emph{\textbf{reasonable coarse-grained hirability decisions are possible even when accurate OCEAN estimates are available to the AHA in lieu of a multimedia CV}}.


\subsubsection{Explaining visual Predictions}
\begin{figure*}[t]
\centering
\includegraphics[width=\textwidth, height=3.5cm]{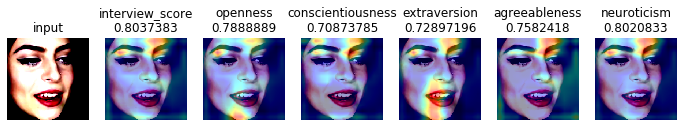}
\includegraphics[width=\textwidth, height=3.5cm]{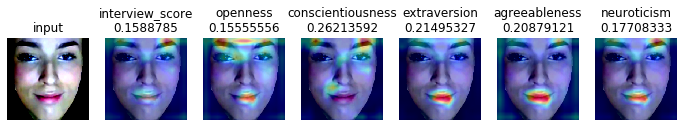}
\vspace{-.6cm}
\caption{Exemplar grad-cam outputs for a person eliciting \emph{high} trait scores (top) and \emph{low} trait scores (bottom). Eyes are the primary cue for eliciting apparent Conscientiousness impressions, while other traits are influenced by holistic facial structure and facial emotions. Best-viewed in color.}\label{fig:gradcam_face} 
\vspace{.1cm}
%
%
\centering
\includegraphics[width=\textwidth, height=3.5cm]{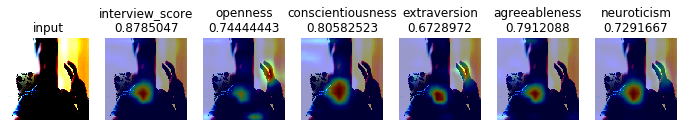}
\includegraphics[width=\textwidth, height=3.5cm]{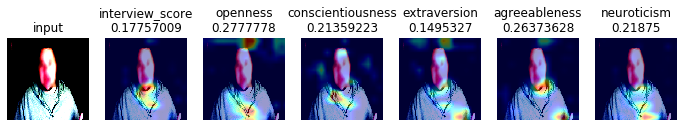}
\vspace{-.6cm}
\caption{Exemplar grad-cam outputs on blurred face portraits for a person eliciting \emph{high} trait scores (top) and \emph{low} trait scores (bottom). Attention maps indicate a focus on the neck region, which determines the relative orientation between the face and body, hand gestures and clothing. Best-viewed in color.}\label{fig:gradcam_blurface} 
\vspace{-.3cm}
\end{figure*}

While the above inferences may be logically derived from experimental results, we explored if any explanations of the visual predictions are possible. Prior works~\cite{Gucluturk2018,escalante2020modeling} show some visual correlates of the IOCEAN traits without \emph{explicitly} showing where their predictive models are looking at. Differently, we employed the Grad-CAM algorithm~\cite{Selvaraju_2019} to \emph{highlight} image regions deemed important for a trait prediction. Using Grad-CAM, gradients of the IOCEAN output neurons are used to get a weighted-sum of the convolutional layer output maps, termed \emph{attention maps} depicting where the network \emph{sees} to accurately predict the trait. We generated activation maps for the IOCEAN traits highlighting important facial and body cues (Figures~\ref{fig:gradcam_face},~\ref{fig:gradcam_blurface}).

Fig.~\ref{fig:gradcam_face} shows Grad-CAM outputs for a \emph{high} and a \emph{low} trait exemplar. One can note that the attention maps relate to the eye and the mouth regions for the IOCEAN traits, which are likely to be of interest to a human interviewer as well. Conscientiousness is one (possible) exception where attention is more localized to the eyes. Conscientiousness is associated with sincerity and uprightness, and is traditionally gauged from eye-movement cues~\cite{Hoppe18}. Conversely, when the face is blurred so as to make the facial cues indecipherable (Fig.~\ref{fig:gradcam_blurface}), the activation maps are focused around the neck region, hand movements and clothing. When the face is represented as a {blob}, the neck region becomes important as it determines the relative orientation between the face and body. \emph{\textbf{These visual explanations cumulatively convey the importance of eye and mouth movements, hand gestures and attire for HP. }}   



\section{Discussion \& Conclusion}
At the outset, the objectives of this work were two-fold: (1) to explicitly and rigorously explore the correlations between \emph{hirability} and the OCEAN \emph{personality traits}, given that this dependence has been exploited earlier in a limited way~\cite{Gucluturk2018,escalante2020modeling}, and (2) to provide \emph{explanations} supporting IOCEAN predictions made by the multimodal behavioral models. Based on the experimental results, we conclude that this work has substantially achieved both objectives. 

With respect to (1), we note that continuous/categorical HP from OCEAN estimates, which are in-turn obtained from audio, visual and verbal behaviors, is more effective than directly predicting from behavioral measures. While this may seem surprising, we believe that this result is only an implication of designing a \emph{simple} HP model with only the OCEAN trait predictors, rather than a `black-box' model with high-dimensional inputs but limited interpretability.  

Regarding (2), we note that all considered modalities and features provide some explanations towards IOCEAN prediction. With respect to \emph{text}, we found that IWs of word stems are highly informative; \eg, use of the word \emph{dead} negatively impacts hiring impressions, and conveys anxiety (indicator of Neuroticism). The words \emph{hobbi} and \emph{fashion} convey a high level of Extraversion. Apparent Conscientiousness is negatively impacted by cuss words, but positively by words relating to well-being. While audio-related explanations are not explicitly presented, we note from Figure~\ref{fig:audio_dur} that IOCEAN predictions saturate beyond 6s time-windows, implying that \emph{tiny} behavioral episodes suffice for reliable trait prediction.          

Visual cues are also highly informative, as confirmed by both quantitative and qualitative results. Quantitative results show that the \emph{eye} and \emph{body} cues achieve IOCEAN prediction comparable to \emph{face} cues. This is a useful result, as processing facial information incapable of revealing identity would assuage candidates' privacy concerns. That body cues can achieve high accuracy on categorical IOCEAN prediction implies that fine-grained behavioral analytics may not be necessary for making coarse-grained decisions. Also, Table~\ref{tab:I_OCEAN_con_video} conveys that coarse hiring decisions are possible solely based on a candidate's OCEAN estimates. Grad-CAM visualizations show the influence of \emph{eye} and \emph{mouth movements}, \emph{hand movements} and \emph{attire} on hirability.
 
Limitations of this study include (a) experiments on only the sampled FICS dataset involving $\approx$ 5K videos, with a clear-cut distinction between \emph{high} and \emph{low}-hirability observations; this was nevertheless design to elicit predictive explanations, and (b) experiments on only the FICS dataset. Future work will focus on validating, extending and generalizing current results via experimentation on multiple datasets.  

\bibliographystyle{ACM-Reference-Format}
\bibliography{sample-base}

%
%
%
%
%
%
%
%

\end{document}